%% file: main.tex
\documentclass{article}
\pdfoutput=1

\usepackage{microtype}
\usepackage{graphicx}
\usepackage{subfigure}
\usepackage{booktabs} % for professional tables

\usepackage{colortbl}
\usepackage{xcolor}  
\usepackage{multirow}
\usepackage{paralist} % for horizontal lists
\usepackage{mdframed} % frame boxes
\usepackage{xurl} % support linebreaks in \url

\usepackage{hyperref}

\usepackage[accepted]{icml2025}

\usepackage{amsmath}
\usepackage{amssymb}
\usepackage{mathtools}
\usepackage{amsthm}

\usepackage[capitalize,noabbrev]{cleveref}

\theoremstyle{plain}
\newtheorem{theorem}{Theorem}[section]

\theoremstyle{definition}

\theoremstyle{remark}

\usepackage[textsize=tiny]{todonotes}

\usepackage{xspace}
\newcommand{\ie}{\textit{i.e.}}
\newcommand{\eg}{\textit{e.g.}}

\newcommand{\method}{\textsc{PowerAttention}\xspace}

\icmltitlerunning{PowerAttention: Exponentially Scaling of Receptive Fields for Effective Sparse Attention}

\begin{document}

\twocolumn[
\icmltitle{PowerAttention: Exponentially Scaling of Receptive Fields for Effective Sparse Attention}



\icmlsetsymbol{equal}{*}

\begin{icmlauthorlist}
\icmlauthor{Lida Chen}{equal,kwlab}
\icmlauthor{Dong Xu}{equal,kwlab}
\icmlauthor{Chenxin An}{hku}
\icmlauthor{Xintao Wang}{kwlab}
\icmlauthor{Yikai Zhang}{kwlab}
\icmlauthor{Jiangjie Chen}{bytedance}
\icmlauthor{Zujie Liang}{antgroup}
\icmlauthor{Feng Wei}{antgroup}
\icmlauthor{Jiaqing Liang}{kwlab}
\icmlauthor{Yanghua Xiao}{kwlab}
\icmlauthor{Wei Wang}{kwlab}
\end{icmlauthorlist}

\icmlaffiliation{kwlab}{Knowledge Works Research Laboratory, School of Computer Science and Technology, Fudan University, Shanghai, China}
\icmlaffiliation{antgroup}{Ant Group, Hangzhou, China}
\icmlaffiliation{bytedance}{ByteDance Seed, Shanghai, China. Part of the work done while at Fudan University.}
\icmlaffiliation{hku}{The University of Hong Kong, Hong Kong, China}

\icmlcorrespondingauthor{Yanghua Xiao}{shawyh@fudan.edu.cn}
\icmlcorrespondingauthor{Wei Wang}{weiwang1@fudan.edu.cn}

\icmlkeywords{Sparse Attention, Long-context Scaling, Large Language Models}

\vskip 0.3in
]




\makeatletter
\renewcommand{\ICML@appearing}{\textit{Preprint.}
}
\makeatother
\printAffiliationsAndNotice{\icmlEqualContribution} % otherwise use the standard text.

\newcommand{\cxan}[1]{\textcolor{blue}{\bf \small [#1 --cxan]}}

\begin{abstract}
\input{000abstract}
\end{abstract}
\section{Introduction}
\label{sec:intro}
\input{010intro}

\section{Related Work}
\label{sec:related}
\input{020related}

\section{Methodology}
\label{sec:method}
\input{030method}

\section{Experiments}
\label{sec:experiments}
\input{040experiments}

% \section{Analysis and Ablation}
% \label{sec:analysis}
% \input{050analysis}

\section{Conclusion}
\label{sec:conclusion}
\input{060conclusion}

% \section*{Acknowledgements}

\section*{Impact Statement}

This paper presents work whose goal is to advance the field of Large Language Models and Attention Mechanisms. There are many potential societal consequences of our work, none of which we feel must be specifically highlighted here.

\bibliography{main}
\bibliographystyle{icml2025}

%%%%%%%%%%%%%%%%%%%%%%%%%%%%%%%%%%%%%%%%%%%%%%%%%%%%%%%%%%%%%%%%%%%%%%%%%%%%%%%
%%%%%%%%%%%%%%%%%%%%%%%%%%%%%%%%%%%%%%%%%%%%%%%%%%%%%%%%%%%%%%%%%%%%%%%%%%%%%%%
% APPENDIX
%%%%%%%%%%%%%%%%%%%%%%%%%%%%%%%%%%%%%%%%%%%%%%%%%%%%%%%%%%%%%%%%%%%%%%%%%%%%%%%
%%%%%%%%%%%%%%%%%%%%%%%%%%%%%%%%%%%%%%%%%%%%%%%%%%%%%%%%%%%%%%%%%%%%%%%%%%%%%%%
\newpage
\appendix
\onecolumn
\input{070appendix}

\end{document}

%% file: 000abstract.tex
Large Language Models (LLMs) face efficiency bottlenecks due to the quadratic complexity of the attention mechanism when processing long contexts. 
Sparse attention methods offer a promising solution, but existing approaches often suffer from incomplete effective context and/or require complex implementation of pipeline.
We present a comprehensive analysis of sparse attention for autoregressive LLMs from the respective of receptive field, recognize the suboptimal nature of existing methods for expanding the receptive field, and introduce \method, a novel sparse attention design that facilitates effective and complete context extension through the theoretical analysis. 
\method achieves exponential receptive field growth in $d$-layer LLMs, allowing each output token to attend to $2^d$ tokens, ensuring completeness and continuity of the receptive field. 
Experiments demonstrate that \method outperforms existing static sparse attention methods by $5\sim 40\%$, especially on tasks demanding long-range dependencies like Passkey Retrieval and RULER, while maintaining a comparable time complexity to sliding window attention. 
Efficiency evaluations further highlight \method's superior speedup in both prefilling and decoding phases compared with dynamic sparse attentions and full attention ($3.0\times$ faster on 128K context), making it a highly effective and user-friendly solution for processing long sequences in LLMs.

%% file: 010intro.tex
\begin{figure}[t]
    \centering
    \includegraphics[width=\columnwidth]{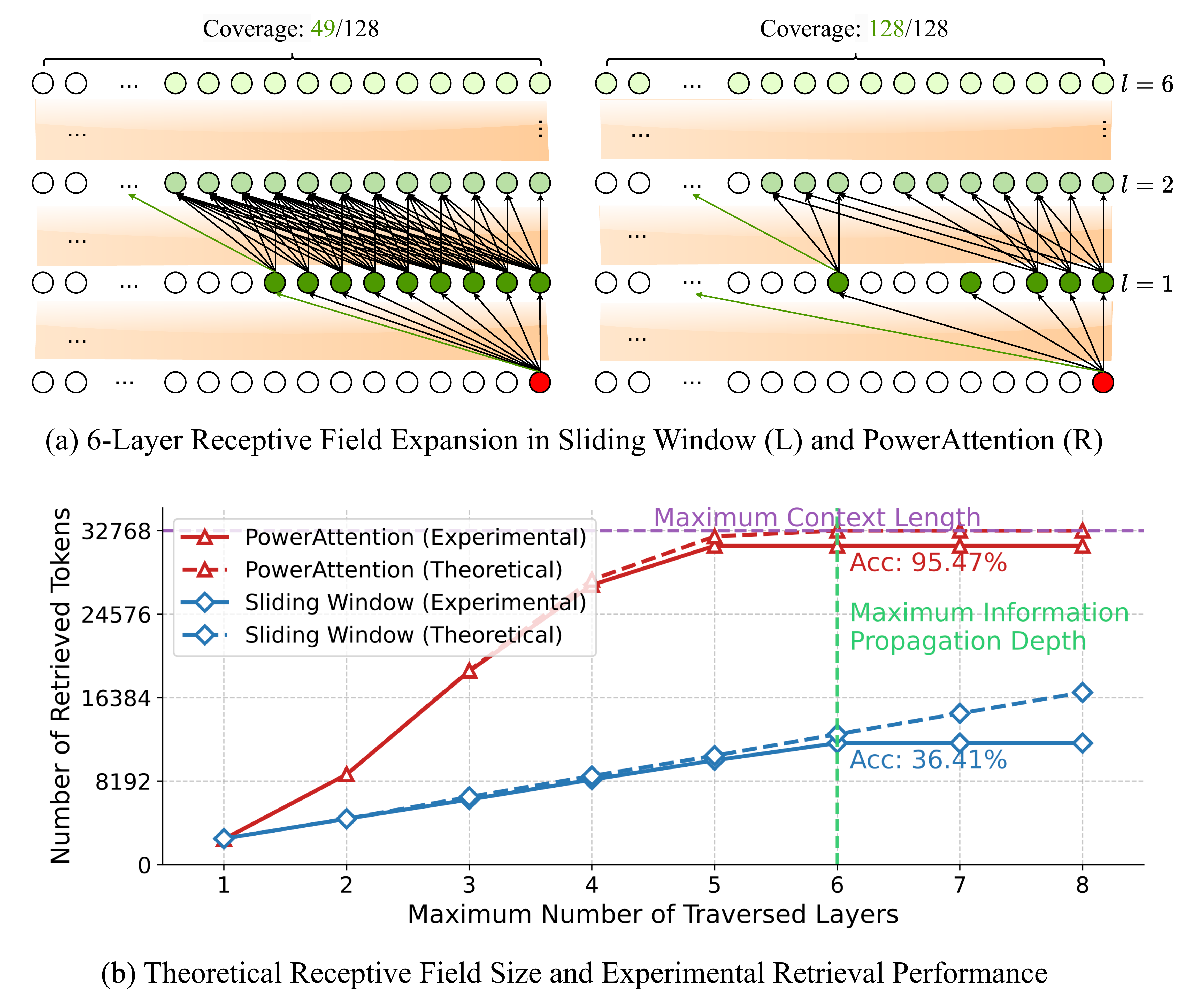}
    \caption{
    Layer-wise receptive field analysis of sparse attention patterns.
    (a) illustrates the information flow across six layers with a simplified 128-block example, while (b) presents the quantitative evaluation on Qwen2-7B with 32K context length.
    The actual token retrieval capability closely matches the theoretical receptive field growth for both patterns.
    Within the maximum information propagation depth, \method's exponential growth in receptive field leads to significantly higher accuracy compared to sliding window's linear expansion.
    Detailed implementation is provided in Appendix \ref{sec:appendix_retrieval_evaluation}.
    }
    \label{fig:intro}
\end{figure}
Large Language Models (LLMs) have demonstrated remarkable capabilities across diverse NLP tasks.
Increasing context length allows LLMs to support more complex applications like long chain-of-thought reasoning~\cite{openai2024o1,qwq2024,deepseekai2025deepseekr1incentivizingreasoningcapability}, agents in complex environments~\cite{Park2023GenerativeAgents,zhou2023webarena,chen2023put}, and long document question answering~\cite{chevalier2024language,wang-etal-2024-leave, leval}.
% The superior performance of Transformer-based LLMs is rooted in the attention mechanism, which enables them to outperform models with other architectures (\ie, Recurrent Neural Networks, or RNNs, and Convolutional Neural Networks, or CNNs) across general tasks[\todo], making Transformer the dominant architecture for LLMs.
However, the quadratic complexity of the attention mechanism poses a significant efficiency bottleneck for Transformer-based LLMs when processing long contexts.

To address the inefficiency of Transformer, recent studies have explored sparse attention~\cite{DBLP:conf/emnlp/CorreiaNM19,beltagy2020longformerlongdocumenttransformer,DBLP:journals/tacl/RoySVG21,9896137,DBLP:conf/nips/AnagnostidisPBN23,jiangminference}, which reduces computational complexity by restricting each token to attend to only a fixed number of tokens instead of the full sequence.
The \textit{static} pattern uses a pre-defined attention mask such as the classic sliding window attention, while the \textit{dynamic} pattern requires the model to be trained with full attention and to update the defined attention mask at inference stage, such as MInference~\cite{jiangminference}.
Dynamic patterns usually achieves better performance in downstream tasks, but the static counterpart features efficiency optimization in training stage and can better handle new tokens during decoding.

However, both the two mainstream sparse attention methods have predominantly relied on intuitive heuristics and experimental results, lacking theoretical analysis to explain their effectiveness.
In this paper, we address this critical gap by presenting a novel comprehensive analysis of sparse attention methods for autoregressive LLMs, providing new insights for designing efficient attention for the future.
Our analysis starts from the information flow across LLM layers.
Consider how information flows within an LLM: at each layer, a token receives information from other tokens it can attend to via self-attention and propagates this aggregated information to subsequent layers.
To analyze this process systematically, we introduce the concept of \textit{\textbf{model receptive field}}, defined as the maximum set of context tokens that the model can utilize during output generation, and model it as a Directed Acyclic Graph (DAG) where different static sparse attention patterns correspond to different edge sets.
Although different sparse patterns with the same sparsity result in identical single-layer receptive field sizes, well-designed patterns can achieve much larger effective receptive fields across multiple layers through efficient information propagation (Figure~\ref{fig:intro}(a)).

Based on this analysis framework, we identify two critical limitations of existing static sparse attention designs that prevent them from achieving optimal receptive fields:
(1) information from tokens at certain positions cannot be retrieved by the final output,
and (2) they exhibit low efficiency in expanding the receptive field layer by layer, as demonstrated by sliding window's linear growth in Figure~\ref{fig:intro}(b).
Based on these insights, we propose \method, a novel sparse attention pattern that can achieve an effective balance between efficiency and performance, both theoretically and experimentally (Figure~\ref{fig:intro}(b)).
Specifically, by calculating attention between tokens at power-of-2 distances, \method achieves exponential expansion of the receptive field across layers while requiring only $O(\log n)$ tokens to ensure the receptive field covers the entire sequence, demonstrating significant potential for ultra-long sequences and high-sparsity scenarios.

We conduct comprehensive experiments to evaluate both the model performance and efficiency of existing static sparse attention methods and \method.
On long-range dependency tasks like Passkey Retrieval and RULER, \method significantly outperforms other static sparse attention methods.
In terms of efficiency, static sparse attention methods with the same sparsity show similar performance, and outperform both full attention and dynamic sparse attention methods like MInference in both prefilling and decoding phases by $1.2\sim30\times$.

In summary, our contributions are:
\begin{itemize}
    \item We establish an analysis framework for studying static sparse attention patterns in autoregressive LLMs, which explains why certain patterns are effective.
    \item We design a novel static sparse attention pattern, \method, that achieves the best balance between efficiency and performance, both theoretically and experimentally.
    \item We conduct extensive experiments demonstrating that \method achieves superior performance compared to existing static sparse attention methods while maintaining state-of-the-art efficiency.
\end{itemize}

%% file: 020related.tex
\paragraph{Dynamic Sparse Attention}
It has been widely observed that attention patterns are often highly sparse~\cite{9896137}, allowing certain correlation computations between tokens to be omitted without significantly degrading the model performance. 
Dynamic sparse attention mechanism predicts the necessary sparse pattern based on the input context and relevant information, which focuses on either prefilling~\cite{DBLP:journals/tacl/RoySVG21,DBLP:conf/asplos/Qu0TCD022,9896137,DBLP:conf/icml/RibarCHBLO24,jiangminference,gao2024seerattentionlearningintrinsicsparse}, adapting the overall attention pattern to the entire input sequence, or focuses on decoding, dynamically evicting~\cite{DBLP:conf/nips/AnagnostidisPBN23,DBLP:conf/nips/Zhang00CZC0TRBW23,DBLP:conf/iclr/Ge0LZ0024,DBLP:conf/icml/ZhangZYXLPJ24,hooper2024squeezedattentionacceleratinglong} or selecting~\cite{DBLP:conf/icml/TangZZXKH24,li2024snapkvllmknowslooking,DBLP:conf/emnlp/KimSCC24,yang2024tidaldecodefastaccuratellm} tokens from the offloaded KV-Cache in each iteration. 
While this dynamic nature allows for greater expressiveness, it also introduces extraneous complexity, both in implementation and computation. 
For instance, dynamic prefilling methods, requiring $O(N)$ time complexity (a whole line scan) at each step, will incur an additional quadratic time overhead to decode $N$ tokens, which prevents them from providing a substantial speedup during the decoding stage~\cite{jiangminference}.

\paragraph{Static Sparse Attention}
Static sparse attention, by contrast, represents a more straightforward design, in which a predefined masking pattern is applied consistently throughout the inference process. 
A common static sparse attention consists of some initial tokens at the beginning (\textit{sink tokens}) and a fixed sliding window (\textit{local windows}). 
This pattern has proven effective in generating fluent outputs with low perplexity~\cite{DBLP:conf/iclr/XiaoTCHL24,DBLP:conf/naacl/HanWPX0JW24}. 
Earlier work includes fixed strided patterns~\cite{child2019generatinglongsequencessparse,DBLP:conf/icml/ShiGRXLLK21}, dilated patterns~\cite{ding2023longnetscalingtransformers1000000000}, or a mix of both~\cite{beltagy2020longformerlongdocumenttransformer,DBLP:conf/nips/ZaheerGDAAOPRWY20}. 
However, sliding window-based solutions fail to effectively leverage information from long contexts. 
Specifically, in sliding window attention with a window size of $W$, each position within the hidden state attends exclusively to the $W$ preceding positions. 
This inherent locality implies that the range of context that sliding window attention can capture (which we refer to as \emph{receptive field} later) expands linearly with the number of layers. 
Our method, \method, also falls into this category of mechanisms. 
Recognizing the limitation, it aims to optimally extend the effective context length while minimizing the overhead of additional computations.
 
\paragraph{Alternative Architecture for Transformer}
Alternative architectures have been proposed to replace modules of the traditional transformer, including state-space model~\cite{DBLP:conf/icml/PoliMNFDBBER23,peng2023rwkvreinventingrnnstransformer,sun2023retentivenetworksuccessortransformer,gu2024mambalineartimesequencemodeling,de2024griffinmixinggatedlinear,DBLP:conf/iclr/LiCZCD23}, linear attention~\cite{DBLP:conf/icml/KatharopoulosV020,feng2024attentionrnn} and long-term memory~\cite{behrouz2024titanslearningmemorizetest}.
While these methods have demonstrated superior language modeling capabilities on certain tasks, they have not yet seen widespread adoption in real-world applications.

%% file: 030method.tex
\begin{figure*}[t]
    \centering
    \includegraphics[width=\linewidth]{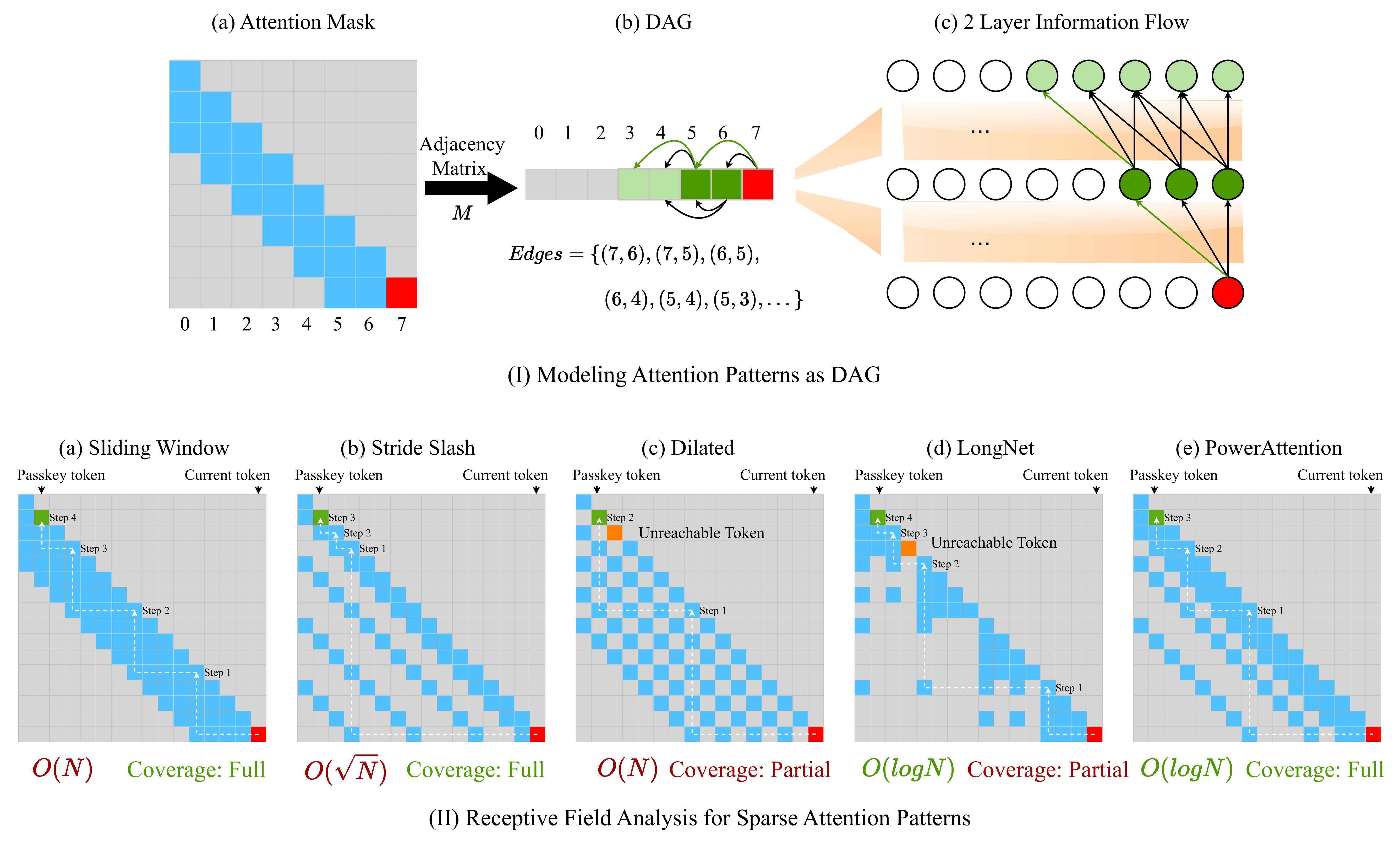}
    \caption{
    (I) \textbf{Modeling Attention Patterns as DAG}: 
    the attention mask serves as the adjacency matrix of a DAG, where edges represent token connections across layers, and the shortest path length indicates the minimum number of layers required for information flow between tokens.
    (II) \textbf{Receptive Field Analysis for Sparse Attention Patterns}:
     white lines show the shortest path to reach passkey tokens, with path length complexity $O(f(N))$ for distance $N$ and coverage indicating token accessibility.
    }
    \label{fig:method}
\end{figure*}
\subsection{Problem Formulation}\label{subsec:problem_formulation}

%图1:attention mask作为临接矩阵->单层DAG边集（边集(1,2),(2,3),...）->2层展开
LLM's decoder layer leverages the attention mechanism to incorporate contextual information into the token generation process, as formally expressed by the following equation:

\begin{equation}
    \text{o}_i = \sum_{j \in \mathcal{A}_i} \text{softmax}\left(\frac{q_i k_j^T}{\sqrt{d_k}}\right)v_j
\end{equation}

where $\text{o}_i$ is computed as a weighted sum of value vectors ($v_j$) from the attended tokens ($j \in \mathcal{A}_i$), with weights determined by similarity scores between $q_i$ and $k_j$.
We define the \textit{receptive field} as the set of all tokens that can influence a given token's representation.
Specifically, the single-layer receptive field of token $i$ is $\mathcal{A}_i$, which directly influences the $\text{o}_i$ within the current layer.
While full attention allows every token to attend to all previous tokens ($\mathcal{A}_i = \{j \in \mathbb{Z}^*| j \leq i\}$), sparse attention strategically limits attention to a subset of tokens to reduce computational costs.
However, this restriction poses the risk of omitting crucial information which lies outside the receptive field.

We formulate token selection in sparse attention as the problem of finding an optimal edge set in a graph, where nodes represent tokens at specific positions.
Sparse attention masks can be naturally interpreted as adjacency matrices, as illustrated in Figure~\ref{fig:method}.
Since modern LLMs adopt autoregressive mechanism, the graph is a directed acyclic graph (DAG) where each token can attend only to earlier ones.
Within a single layer, a given token's receptive field consists of all its successor nodes in the graph.

Although different sparse patterns at comparable sparsity result in similar out-degree of nodes (\ie, the size of single-layer receptive field), well-designed patterns can achieve larger effective receptive fields across multiple layers.
Consider the internal information flow within an LLM during a single forward pass: at layer $l$, the token representation at position $i$ receives information from tokens within its single-layer receptive field via self-attention, which is then propagated through the feed-forward layer to the next layer.
Through this process, $\text{o}_i$ at layer $l$ effectively relays information from previous tokens, serving as an intermediate node that propagates aggregated information to tokens that attend to $i$ in subsequent layers.
For instance, in a two-layer scenario, when token $x$ attends to $y$ in the second layer, $y$'s representation already encodes first-layer information, thereby expanding $x$'s receptive field effectively.
Thus, in multi-layer LLMs, the receptive field of token $x$ extends beyond immediate successors to encompass all DAG-accessible nodes originating from $x$.
We conduct an empirical study on information flow in Section \ref{subsec:probing}.

Therefore, under the constraint of preserving the computational efficiency, we can reformulate the problem of finding the optimal sparse attention design as \textbf{finding an edge set in the DAG that maximizes node reachability in $l$ steps under fixed maximum out-degree constraints} ($l$ represents the number of model layers).
For nodes beyond a distance of $l$, the model theoretically cannot access their information when predicting the next token.
Consequently, if these tokens contain key information, the model performance will degrade significantly.

\subsection{Limitations of Existing Sparse Attention}
We analyze several static sparse attention designs:
(1) Sliding window~\cite{DBLP:conf/iclr/XiaoTCHL24,DBLP:conf/naacl/HanWPX0JW24}, which incorporates attention sink tokens from the sequence start in addition to the local window;
(2) Stride slash attention~\cite{child2019generatinglongsequencessparse}, which places slash tokens at equal intervals across the context length, beyond the local window and sink tokens;
(3) Dilated attention~\cite{beltagy2020longformerlongdocumenttransformer}, which employs dilated local windows;
(4) LongNet~\cite{ding2023longnetscalingtransformers1000000000}, which constructs the attention mask by overlaying multiple masks with geometrically increasing block sizes and dilated intervals;

We analyze the shortest path from the last token to reach a passkey token in different attention designs.
As shown in Figure~\ref{fig:method}, in sliding window attention, each token can reach the farthest token within its window until the passkey token appears in the window. To reach a token at distance N, it requires $O(N)$ layers.
Under stride slash attention, a token first reaches the nearest slash token to its target, then iteratively reaches the farthest token within each window until the passkey token appears.
With strategically placed slash tokens, reaching a token at distance N only requires $O\left(\sqrt{N}\right)$ layers.
Both dilated attention and LongNet have unreachable tokens, making them impossible to retrieve passkeys at certain positions.
In dilated attention, all tokens at distances $2k+1$ from the current token are unreachable. 
Despite having a window twice as large as sliding window at the same sparsity, it can only access 50\% of the tokens.
LongNet requires $O(logN)$ layers to reach a token at distance N, but cannot access certain tokens, such as the last token in each segment.
Therefore, existing methods often fail to achieve both fast expansion of the receptive field and complete token coverage.

\subsection{PowerAttention}
Based on our modeling of sparse attention, we propose \method, a sparse attention design that exponentially expands the receptive field. 
Our edge set construction ensures that in a DAG, any node can reach all nodes within a distance of $n$ in at most $\log n$ steps, while maintaining a maximum out-degree of $\log n$. 
This is achieved by connecting each node only to nodes whose index differences are powers of 2, which is transformed to a sparse pattern where each token attends only to positions at power-of-2 distances.

Under our pattern, we guarantee that the receptive field grows exponentially with the maximum distance $d$, while capturing information from all tokens within a distance of $2^d$. 
The theoretical proof of this property is provided in Appendix~\ref{sec:appendix_proof}. 
As for implementation, \method introduces no additional computational overhead.
We present its pseudo-code implementation in Algorithm~\ref{alg:power_attention}.

\begin{algorithm}[htbp]
    \caption{\method in Python-like pseudo-code}
    \label{alg:power_attention}
    \includegraphics[width=\linewidth]{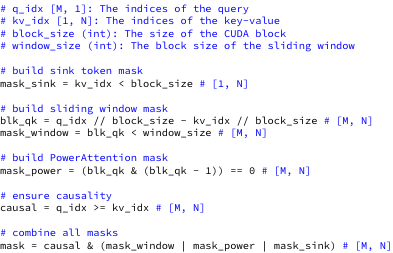}
    % \begin{algorithmic}[1]
    %     \REQUIRE Query index q\_idx, Key-Value index kv\_idx, Batch size b, Head number h, Block size B, Window size W
    %     \ENSURE Attention mask for the position (q\_idx, kv\_idx)
        
    %     \STATE $\text{blk\_qk} \gets \lfloor\text{q\_idx}/\text{B}\rfloor - \lfloor\text{kv\_idx}/\text{B}\rfloor$
        
    %     \STATE $\text{mask\_window} \gets \text{blk\_qk} < \text{W}$
        
    %     \STATE $\text{mask\_power} \gets (\text{blk\_qk} \& (\text{blk\_qk} - 1)) == 0$ ~// Power-of-2 attention

    %     \STATE $\text{mask\_sink} \gets \text{kv\_idx} < \text{B}$

    %     \STATE $\text{mask\_casual} \gets \text{q\_idx} \geq \text{kv\_idx}$
        
    %     \RETURN $\text{mask\_casual} \land (\text{mask\_window} \lor \text{mask\_power} \lor \text{mask\_sink})$
    % \end{algorithmic}
\end{algorithm}

%% file: 040experiments.tex
\input{Tables/ppl}

\begin{figure*}[t]
    \centering
    \includegraphics[width=\linewidth]{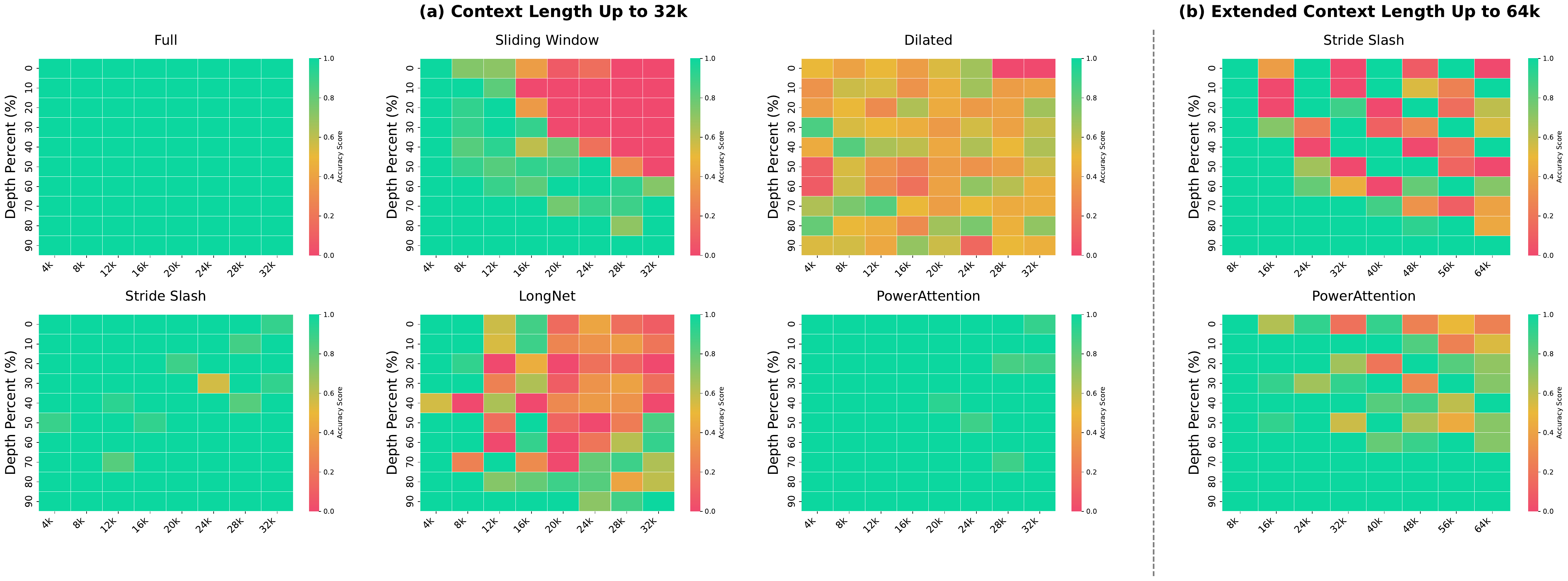}
    \caption{
    Results on passkey retrieval with different attention patterns: \textbf{(a)} evaluation on context lengths up to 32k, and \textbf{(b)} comparison between stride slash attention and \method on extended context lengths up to 64k.}
    \label{fig:passkey}
\end{figure*}
We evaluate the performance of existing sparse attention designs and \method on LLMs in terms of both accuracy and efficiency.
For accuracy, we assess model performance on language modeling, synthetic retrieval tasks, and long context benchmarks.
For efficiency, we measure model performance during both the prefill and decoding phases.

\subsection{Setting}
\paragraph{Implementation}
We use Qwen2-7B~\cite{yang2024qwen2technicalreport}, a pretrained model with 32K context length, as our base model.
To better adapt the model to sparse attention patterns, we conduct continued pre-training on processed SlimPajama corpus~\cite{cerebras2023slimpajama} for 1B tokens.
To effectively train the model with sparse attention patterns, we fine-tune the model on ChatQA 2~\cite{xu2024chatqa2bridginggap} data before evaluating on long context benchmarks.
This ensures the training data contains dependencies beyond the local window, providing natural supervision signals for the model to learn and utilize its full receptive field.
For hardware efficiency, we implement sparse attention using 256-token blocks to align with GPU compute cores' memory access pattern.
To ensure a fair comparison across designs, we maintain consistent sparsity levels by fixing the maximum number of tokens each position can attend to, and employ the consistent patterns during both prefilling and decoding phases.

\input{Tables/ruler}

\paragraph{Baseline}
We evaluate the static sparse attention designs analyzed in Section~\ref{sec:method}, including sliding window attention, stride slash attention, dilated attention, LongNet, and \method.
The specific configurations are as follows:
For sliding window attention, we use a 9-block local window and 1 block of sink tokens.
For stride slash attention, we configure a 6-block local window, 1 block of sink tokens, and 3 blocks of slash tokens.
For dilated attention, we set the local window size to 20 blocks with a dilation rate of 1 block.
For LongNet, we use segment lengths w = {8, 16, 32, 64, 128} blocks with corresponding dilation ratios r = {1, 2, 4, 8, 16}.
For \method, we employ a 5-block local window, 1 block of initial tokens, and 4 additional blocks of slash tokens distributed at power-of-2 intervals.
See Appendix~\ref{sec:appendix_baseline_impl} for more implementation details.

\subsection{Long Context Language Modeling}
We first evaluate the language modeling perplexity of different static sparse attention methods on the PG19 test set~\cite{raecompressive2019}.
The evaluation is conducted across four different sequence lengths, as shown in~\ref{tab:perplexity-single-col}.
All sparse attention methods maintain low perplexity scores up to 32K context length, demonstrating their ability to preserve strong language modeling capabilities while reducing computational costs.
Notably, despite using the same number of attended tokens, different sparse attention designs vary in performance.
StreamingLLM, \method, and stride slash attention achieve lower perplexity, while dilated attention and LongNet perform slightly worse.
We attribute this gap to the discontinuous receptive fields in the latter two, which may hinder effective language modeling.

\begin{figure*}[t]
    \centering
    \subfigure[End-to-end inference latency of different attention methods with 16K$\sim$128K input context and 1024 decoding steps.]{
        \includegraphics[width=0.4\textwidth]{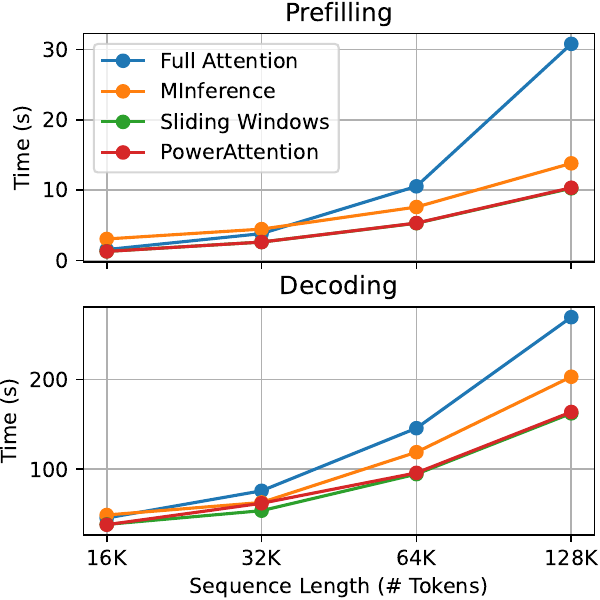}
        \label{fig:efficiency_endtoend}
    }
    \hskip 0.1in
    \subfigure[Time consumption for each forward pass of the attention kernel under different methods with 16K$\sim$128K input context.]{
        \includegraphics[width=0.4\textwidth]{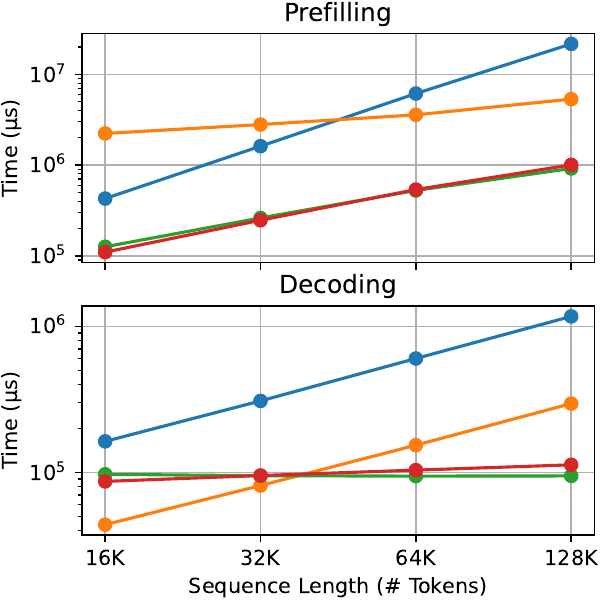}
        \label{fig:efficiency_kernel}
    }
    \caption{
    Efficiency evaluation results on Qwen2-7B with a NVIDIA A800 GPU.
    }
\end{figure*}

\subsection{Retrieval-Based Evaluation}\label{subsec:passkey_retrieval}

While local information suffices for low perplexity in language modeling, it inadequately assesses a model's ability to capture key information across the entire context.
To evaluate how different static sparse attention designs affect the model's ability to capture context-wide information through their receptive fields, we evaluate them on the passkey retrieval~\cite{mohtashami2023randomaccess} task.
% To complete this task, the model must capture the passkey information that can be located anywhere in the context to generate the correct response.
This requires the model's receptive field to seamlessly cover all positions and efficiently span the entire sequence within limited layers.
To isolate training data effects, we fine-tune LLMs on the same synthetic passkey retrieval dataset.
We also employ curriculum learning, gradually increasing sequence lengths from 4K to 32K with 200 steps per stage.

Figure~\ref{fig:passkey} demonstrates that sparse attention performance aligns with our theoretical analysis of receptive field:
sliding window performs well up to 12K tokens but degrades significantly at 16K and 32K lengths, failing to retrieve passkeys from initial 16K tokens at 32K length.
With a window size of approximately 2K tokens, we estimate that information strength decays to near zero after propagating through 6 layers.
Dilated attention captures only $\sim50\%$ input across all lengths, as its 5K window covers the 30K sequence within 6 layers but lacks adjacent-block aggregation in the last block.
LongNet similarly suffers from gaps in its receptive field, failing to capture information at specific positions.
Both stride slash attention and \method achieve full-sequence coverage within 6 layers, enabling successful 32K passkey retrieval.
Notably, further evaluation at 64K sequence length shows that \method's exponential receptive field expansion achieves better performance than stride slash's quadratic approach, demonstrating the benefits of faster receptive field growth for ultra-long sequence processing.

\subsection{Long Context Benchmark Evaluation}
To further validate that \method is more effective for long sequence processing, we evaluate all baseline methods on the widely-used RULER benchmark~\cite{hsieh2024rulerwhatsrealcontext}.
RULER is a challenging dataset consisting of 14 sub-tasks that assess models' effective context utilization across different difficulty levels.
We adopt a practical hybrid architecture~\cite{stripedhyena,lieber2024jambahybridtransformermambalanguage,yang2025parallelizinglineartransformersdelta} that retains some full attention layers while varying sparse attention patterns in the remaining layers.
To maximize the continuity of sparse attention layers, we keep 2 full attention layers for every 7 layers.

Table~\ref{tab:ruler} presents the performance comparison on RULER benchmark.
Among all baseline methods, \method demonstrates superior performance by consistently achieving the highest scores across four different sequence lengths.
Meanwhile, stride slash attention and LongNet show comparable performance, ranking below \method but above other baselines.
These results align with our findings from the passkey retrieval task, demonstrating that well-designed sparse attention patterns can achieve larger receptive fields and better handle long sequences.
All sparse attention methods show a noticeable performance gap compared to full attention, which can be attributed to the high sparsity ratio (up to 94\% in sparse attention layers) we employ to simulate ultra-long sequence performance at 32K context length.
Notably, the widely adopted sliding window pattern achieves the lowest accuracy.
Given its prevalence in attention optimization methods~\cite{DBLP:conf/icml/AroraEZTA0RR24}, replacing sliding window with alternative designs may potentially yield further performance improvement.

\subsection{Efficiency Evaluation}
To verify the inference efficiency of \method, we compare its latency against other methods.
We exclude stride slash attention from this evaluation as it obviously has the same computational cost as \method.

\paragraph{End-to-end Latency} 
\method achieves the highest speedup compared with three baseline methods: full attention, sliding window, and MInference. 
Notably, MInference is claimed be more efficient with FlashAttention-based full attention during the decoding phase in the original paper, and we adopted this suggestion, applying MInference solely during the prefilling stage. 
Figure \ref{fig:efficiency_endtoend} shows the latency of different methods. 

At a context length of 128K, \method delivers speedups of 3.0$\times$ and 1.3$\times$ over full attention and MInference in the prefilling phase, respectively. 
In the decoding phase, \method also demonstrates improvements, taking only 58\% and 80\% of the time required by these two methods. 
Consequently, \method offers a user experience that is nearly equivalent to that of sliding window attention.

\begin{figure*}[htbp]
    \centering
    \includegraphics[width=\textwidth]{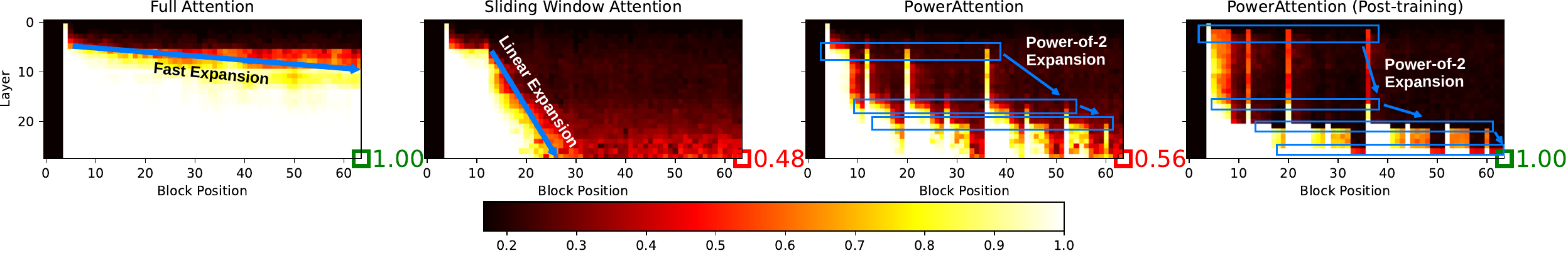}
    \caption{Information flow probing result for various attention mechanisms on the 28-layer Qwen2-7B with 16K context length.
    The sequence is divided into 64 blocks (0 is at the beginning and 63 is at the end), each of which contains 256 tokens, as detailed in Appendix \ref{sec:appendix_probing}.
    Each pixel represents the strength of passkey information at a specific layer and block position; a brighter pixel indicates a higher possibility of extracting passkey information from that position.
    The classification accuracy of the final token block in the last layer is highlighted, as it directly determines the output token.
    }
    \label{fig:probing_result}
\end{figure*}

\paragraph{Kernel Speedup}
To better illustrate the differences in time complexity among various attention methods, Figure \ref{fig:efficiency_kernel} outlines the time taken for each attention forward pass.
Due to \method's inherent $O(N \log_2 N)$ time complexity, its growth rate is nearly as gradual as that of sliding window attention, which has linear time complexity.
At a context length of 128K, \method is 5.3$\times$ faster than MInference and 21.6$\times$ faster than full attention in terms of kernel overhead.
In longer input contexts, \method's attention inference overhead is expected to be superior to these baseline methods.

\subsection{Probing of Information Flow}\label{subsec:probing}

We also investigate LLMs to address the following questions of interest:
\begin{inparaenum}[(1)]
  \item whether the inter-layer information flow, which is the foundation of the receptive field expansion discussed in Section \ref{subsec:problem_formulation}, actually exists in the LLMs, and
  \item whether our proposed training method enable LLMs to leverage this mechanism more effectively.
\end{inparaenum}

To interpret the hidden states propagated within the model, we employ a simple yet effective probing method: linear classifiers.
For convenience, we design a straightforward passkey retrieval task, in which the model is given a long sequence input which includes a specific passkey (\eg, ``\textit{Passkey is apple.}") at a fixed position. 
The passkey is sampled from 6 words (\ie, \textit{apple}, \textit{grape}, etc.), and the remainder of the context is filled with grammatically correct, randomly generated sentences to a length of $N=16\text{K}$. 
The model is asked to identify and extract the passkey from the context in the task.
For each layer of the model, a logistic classifier is trained at each token position to map the hidden state to the corresponding target passkey.
Finally, we collect the prediction accuracies of all classifiers as an indicator of whether passkey information is present at that location; if the state at a position is easily mapped to the corresponding label, it obviously includes the information, and vice versa.
Appendix \ref{sec:appendix_probing} discusses our probing process in details and presents additional probing results for alternative methods post-training.

The probing results are shown in Figure \ref{fig:probing_result}, from which we could conclude that:

\textbf{Inter-layer information flow inherently exists within LLMs.}
In the early layers of the unmodified model with full attention, passkey information is predominantly localized to token positions near the passkey.
As inference progresses through the layers, the information range gradually expands forward throughout the hidden state.
This suggests that even though full attention theoretically allows it to attend to any position in a single step, the attention heads still exhibit a degree of spatial locality.
Specifically, they not only retrieve the original information at the passkey's location but also attend to and aggregate information from neighboring positions where earlier layers have accumulated relevant information.
In sparse attentions, this phenomenon is even more evident: the receptive field of sliding window attention expands progressively across layers at a linear rate, and the receptive field of \method, in contrast, exhibits phase transition-like jumps across layers, enabling information to be flowed in discrete leaps to power-differentiated positions at specific layers.
This demonstrates that inter-layer information flow is an inherent mechanism of LLMs.

\textbf{\method effectively enhances the information flow mechanism.} 
Neither the sliding window attention nor the untrained \method effectively retrieves the correct passkey, although the latter shows a slight advantage.
However, after continued pretraining and finetuning, the model's information flow mechanism improves significantly, with the classification accuracy of the last token in the final layer increasing by 44\% to achieve 100\%.
Visually, the post-training information flow image also exhibits clearer and more focused boundary.
Combining our primary experimental results, we can attribute the post-training performance improvement to the enhancement of this mechanism.

%% file: Tables/ppl.tex
\begin{table}[t]
\centering
\caption{Perplexity of different static sparse attention methods on PG19. 
Each static sparse attention pattern achieves a sparsity ratio of 0.94.
}
\vskip 0.15in
\begin{tabular}{l r r r r}
\toprule
 & \multicolumn{4}{c}{\textbf{PG19}} \\
\cmidrule(lr){2-5}
\textbf{Method} & 4k & 8k & 16k & 32k\\

\midrule
\rowcolor{gray!10} \multicolumn{5}{c}{\textit{\textbf{Full Attention}}} \\
Vanilla        & 9.77  & 9.72  & 9.60 & 9.42 \\
\midrule
\rowcolor{gray!10} \multicolumn{5}{c}{\textit{\textbf{Sparse Attention}}} \\
Sliding Window   & 9.94  & 9.99  & 9.99  & 9.97  \\
Dilated        & 10.74 & 10.72 & 10.64 & 10.58 \\
Stride Slash   & 10.01 & 10.11 & 10.07 & 10.03 \\
LongNet        & 10.14 & 10.28 & 10.30 & 10.31 \\
\method & 10.03 & 10.08 & 10.05 & 10.00 \\
\bottomrule
\end{tabular}

\label{tab:perplexity-single-col}
\end{table}

%% file: Tables/ruler.tex
\begin{table*}[ht]
    \fontsize{15}{20}\selectfont
    \setlength{\tabcolsep}{3pt}
    \centering
    \caption{
    Performance comparison of different static sparse attention patterns on RULER benchmark across different context lengths (4k-32k).
    The RULER benchmark consists of 13 tasks categorized into Needle-in-a-Haystack (NIAH), Variable Tracing (VT), Aggregation, and Question Answering (QA).
    Results show average scores for each category and overall performance across all 13 tasks.
    }
    \vskip 0.15in
    \resizebox{\textwidth}{!}{
    \begin{tabular}{l ccccc ccccc ccccc ccccc ccccc}
    \toprule
    
    \multirow{2}{*}{\textbf{Method}} & \multicolumn{5}{c}{\textbf{4k}} & \multicolumn{5}{c}{\textbf{8k}} & \multicolumn{5}{c}{\textbf{16k}} & \multicolumn{5}{c}{\textbf{32k}} \\
    \cmidrule(lr){2-6}\cmidrule(lr){7-11}\cmidrule(lr){12-16}\cmidrule(lr){17-21}
     & NIAH & VT & Agg. & QA & \textbf{Avg.} & NIAH & VT & Agg. & QA & \textbf{Avg.} & NIAH & VT & Agg. & QA & \textbf{Avg.} & NIAH & VT & Agg. & QA & \textbf{Avg.}   \\
     
    \midrule 
    \rowcolor{gray!10} \multicolumn{21}{c}{\textit{\textbf{Full Attention}}} \\
    Vanilla & 95.69 & 97.60 & 93.48 & 71.50 & 91.77 &95.72 & 91.80 & 68.45 & 64.00 & 86.34 &95.34 & 89.00 & 59.85 & 62.00 & 84.27 &91.38 & 84.60 & 53.75 & 53.50 & 79.24 \\
    
    \midrule
    \rowcolor{gray!10} \multicolumn{21}{c}{\textit{\textbf{Sparse Attention}}} \\

    Sliding Window & 91.75 & 40.40 & 24.61 & 57.50 & 72.20 &74.47 & 38.40 & 17.02 & 44.00 & 58.17 &59.53 & 22.00 & 29.50 & 42.50 & 49.40 &44.88 & 10.60 & 19.14 & 23.00 & 34.91 \\
    Stride Slash & 87.22 & 36.40 & 37.46 & 61.50 & 71.70 &74.25 & 40.20 & 37.34 & 45.50 & 61.53 &65.06 & 22.20 & 40.21 & 43.00 & 54.55 &54.12 & 14.60 & 35.66 & 33.50 & 45.07 \\
    Dilated & 87.91 & 38.20 & 31.66 & 54.50 & 70.29 &85.56 & 4.40 & 20.15 & 48.50 & 63.55 &71.28 & 0.20 & 20.48 & 49.00 & 54.57 &61.03 & 1.20 & 13.70 & 36.50 & 45.37 \\
    LongNet & 84.44 & 43.00 & 35.20 & 65.50 & 70.76 &75.69 & 24.00 & 27.73 & 48.50 & 60.15 &65.00 & 15.60 & 23.02 & 45.00 & 51.66 &51.41 & 9.80 & 14.15 & 31.50 & 39.41 \\
    \method & 90.00 & 26.80 & 36.70 & 60.50 & \textbf{72.40} & 81.47 & 30.20 & 34.56 & 46.50 & \textbf{64.93} & 73.75 & 25.00 & 41.84 & 44.50 & \textbf{60.59} & 62.94 & 18.40 & 36.84 & 32.00 & \textbf{50.74} \\
    
    \bottomrule
    \end{tabular}
    }
    \label{tab:ruler}
    % \vspace{-15pt}
    \end{table*}

%% file: 060conclusion.tex
We present \method, a novel sparse attention mechanism that addresses the limitations of existing static and dynamic sparse attention patterns in LLMs. 
Leveraging a theoretically grounded framework, \method achieves exponential receptive field expansion with complete token coverage, enhancing information flow while maintaining computational efficiency.

Our work not only offers a simple yet effective alternative to current static sparse patterns, but also establishes a theoretical foundation for designing future sparse attention mechanisms. 
We believe this advancement will contribute to the development of more efficient and capable LLMs, ultimately facilitating their application in tasks involving extensive contextual dependencies.

%% file: 070appendix.tex
\section{Quantitative Evaluation of Retrieval Performance}\label{sec:appendix_retrieval_evaluation}
This section details the evaluation experiments we conduct to quantify the actual receptive field of different static sparse attention methods whose results are shown in Figure \ref{fig:intro}(b).

We design a passkey retrieval task, akin to the one in Section \ref{subsec:passkey_retrieval}, where the passkey is a string of random digits (\eg, \textit{12345678}). 
The background text is populated with repeated irrelevant sentences and padded to a fixed length of 32K. 
The position of the passkey is uniformly distributed throughout the context. 
The models are tasked with retrieving and outputting the corresponding passkey.
Based on the block-sparse pattern we implement, the input sequence is divided into blocks of 256 tokens. 
Within each block, we ensure that there are at least five samples with their passkeys uniformly distributed in the block's range.

Building upon the discussed theory, we calculate the set of blocks $\mathcal{B}_k$ that the final block can reach within $k$ step for various sparse attention methods ($k=1,2,\cdots$), and the \textit{theoretical accuracy} $\hat{\alpha}_k$ as the ratio of total blocks to $|\mathcal{B}_k|$. 
For instance, $\mathcal{B}_1$ for sliding window attention is $\{1,120,121,122,\cdots,128\}$, thus $\hat{\alpha}_1=\frac{10}{128}\approx 7.82\%$.
To retrieve the passkeys from these samples, the model must achieve successful information propagation within at least $k$ layers.

We collect the evaluation results and compute the \textit{experimental accuracy} $\alpha_k$ for each step $k$ as the ratio of the total number of samples to the number of examples that are successfully retrieved within $\mathcal{B}_k$.
Ideally, $\hat{\alpha}_k$ is the least upper bound of $\alpha_k$.
Figure \ref{fig:intro} illustrates the relationship between $k$ and $\alpha_k,\hat{\alpha}_k$ across different attention methods.

\section{Proof of Exponential Receptive Field Growth}
\label{sec:appendix_proof}
\begin{theorem}
For a directed acyclic graph (DAG) with vertices labeled from 1 to n, let the edge set be 
\[E = \{(i,j) \mid i-j = 2^k, k \in \mathbb{Z}^*\}\]
Then the following properties hold:
\begin{enumerate}
    \item For any vertex $i$, the out-degree of $i$ is less than $\log n$.
    \item For any vertices $i$ and $j$ where $j < i$, the distance from $i$ to $j$ is at most $\log n$.
\end{enumerate}
\end{theorem}

\begin{proof}
We prove both of the properties:

(1) For any edge $(i,j) \in E$, we have $i-j = 2^k$ where $2^k < n$. Therefore, $k < \log n$. 
Since each possible value of $k$ corresponds to at most one outgoing edge from vertex $i$, 
the out-degree of any vertex is bounded by $\log n$.

(2) Consider any vertex pair $i$ and $j$ where $j < i$. Let $d = i-j$ be the difference.
Since $d < n$, the \textbf{binary representation} of $d$ has at most $\log n$ bits, 
and consequently, at most $\log n$ ones.

Let $k_1, k_2, ..., k_m$ denote the positions of ones in the binary representation of $d$.
Then we can say:
\[d = \sum_{t=1}^{m} 2^{k_t}\]

This decomposition naturally induces a path from $i$ to $j$:
\[i \rightarrow (i-2^{k_1}) \rightarrow (i-2^{k_1}-2^{k_2}) \rightarrow \cdots \rightarrow (i-\sum_{t=1}^{m-1} 2^{k_t}) \rightarrow j\]

The length of this path equals the number of ones in the binary representation of $d$,
which is at most $\log n$. Therefore, the distance from $i$ to $j$ is at most $\log n$.
\end{proof}

\section{Implementation Details and Additional Results of Probing Analysis}\label{sec:appendix_probing}
\subsection{Implementation}
\begin{figure}[htbp]
    \centering
    \includegraphics[width=\linewidth]{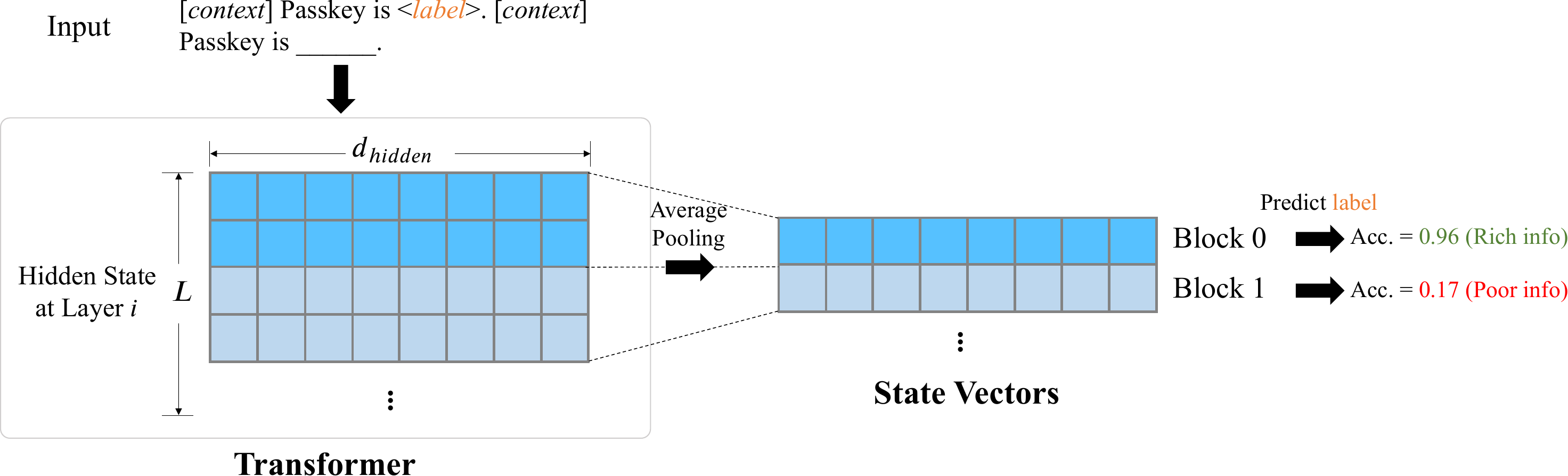}
    \caption{Implementation of our probing analysis on a sequence length of $L$. 
    We employ a linear regression model to evaluate whether a specific block position within a given layer encodes sufficient information of the passkey (\ie, ``Rich info" or ``Poor info").}
    \label{fig:probing}
\end{figure}

Figure \ref{fig:probing} illustrates our approach for probing information flow. 
We feed long sequence retrieval tasks into LLMs, incorporating 6 different passkeys with equal frequencies: \textit{apple}, \textit{banana}, \textit{cherry}, \textit{grape}, \textit{kiwi} and \textit{lemon}; then we collect the hidden states from each layer, which are subsequently average-pooled at evenly spaced intervals, yielding state vectors with a dimensionality of $d_\text{hidden}$. 
For each block within each layer, the state vectors from all samples are gathered and used to train a logistic regression model. 
In other words, with a 28-layer model and 64 sampling positions, we will perform $28\times64=1792$ training runs.

For the classification results, accuracy is directly calculated as the proportion of correctly identified input passkeys across all samples.
With 6 distinct passkeys, if a state vector does not contain any passkey-related information, one would expect a trivial accuracy of $\frac{1}{6}$.

For retrieval tasks, we utilize the following prompt:

\begin{mdframed}
There is an important info hidden inside a lot of irrelevant text. Find it and memorize it. I will quiz you about the important information there.

\textcolor{lightgray}{The abhorrent round combs elevation. The dark roar tabulates event.} [\textit{irrelevant context up to $\sim$1K}] The pass key is \textcolor{orange}{apple}. Remember it. \textcolor{orange}{apple} is the pass key. [\textit{irrelevant context up to $\sim$15K}]

What is the pass key? The pass key is
\end{mdframed}

To ensure consistency and generalizability in decoding and analysis at the same relative position across different samples, we fix the passkey position at the 10\% of the entire context.
We implement all attention patterns using PyTorch's Flex Attention module, and conduct comparative testing on the same task dataset ($N=1200$). 

\subsection{Additional Results on Sliding Window Attention}
\begin{figure}[htbp]
    \centering
    \includegraphics[width=0.8\linewidth]{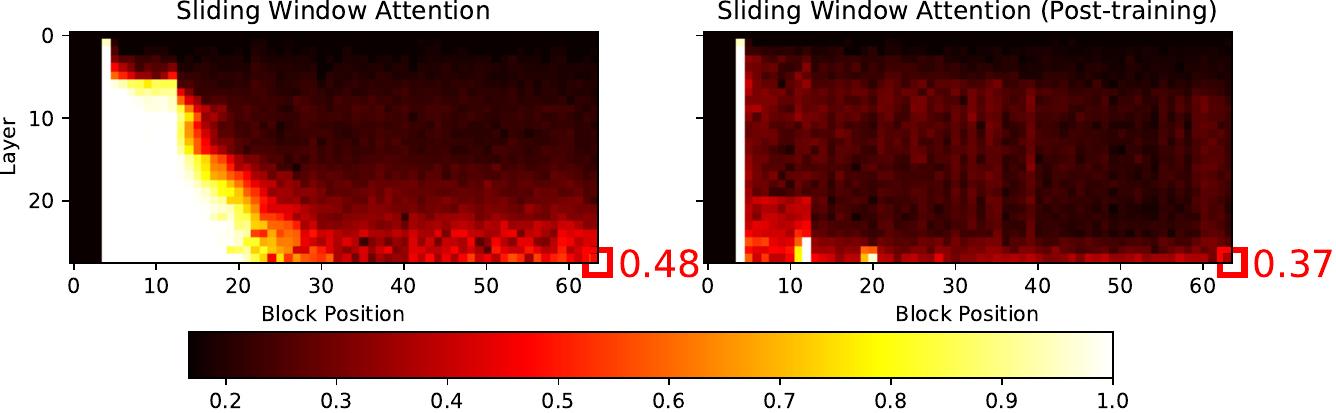}
    \caption{Additional results of information flow probing on sliding window attention.}
    \label{fig:probing_result_extra}
\end{figure}

To maintain consistency in controlled variables, we also conducted probing on the post-trained sliding window attention mechanism, as illustrated in Figure \ref{fig:probing_result_extra}.

Interestingly, the model's performance in information flow degrades post-training, with accuracy declining from $0.48$ to $0.37$. 
We hypothesize this stems from overfitting during the training stage as described in Section \ref{subsec:passkey_retrieval}.
Notably, the model underperforms even on the task to which it overfits (Figure \ref{fig:passkey}).
This observation may highlight fundamental limitations imposed by the inherently restricted receptive field of sliding window attention.

\section{Implementation Details of Baseline Sparse Attention Patterns}
\label{sec:appendix_baseline_impl}
We present the pseudo-code implementations of four baseline sparse attention patterns below.
In our experiments, these patterns are implemented using the FlexAttention~\cite{dong2024flex} library.
Additionally, we provide Triton~\cite{tillet2019triton} implementations combined with RingAttention~\cite{liu2024ringattention} for sequence-parallel training, enabling scaling to longer sequences.

\begin{figure}[htbp]
    \begin{minipage}[t]{0.48\textwidth}
        \raggedright
        \begin{algorithm}[H]
            \caption{Sliding window attention in Python-like pseudo-code}
            \label{alg:sliding_window}
            \includegraphics[width=\linewidth]{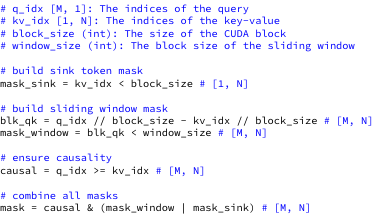}
        \end{algorithm}
    \end{minipage}
    \hfill 
    \begin{minipage}[t]{0.48\textwidth}
        \raggedright
        \begin{algorithm}[H]
            \caption{Stride slash attention in Python-like pseudo-code}
            \label{alg:stride_slash}
            \includegraphics[width=\linewidth]{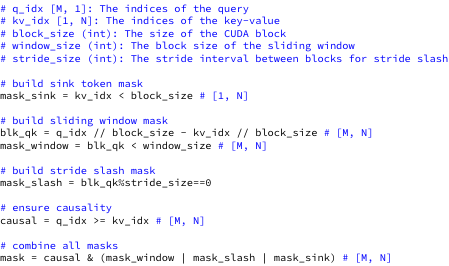}
        \end{algorithm}
    \end{minipage}
\end{figure}

\begin{figure}[htbp]
    \begin{minipage}[t]{0.48\textwidth}
        \raggedright
        \begin{algorithm}[H]
            \caption{Dilated attention in Python-like pseudo-code}
            \label{alg:dilated}
            \includegraphics[width=\linewidth]{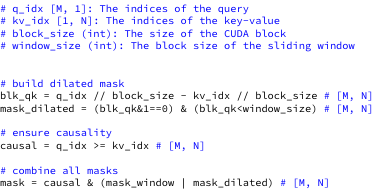}
        \end{algorithm}
    \end{minipage}
    \hfill
    \begin{minipage}[t]{0.48\textwidth}
        \raggedright
        \begin{algorithm}[H]
            \caption{LongNet in Python-like pseudo-code}
            \label{alg:longnet}
            \includegraphics[width=\linewidth]{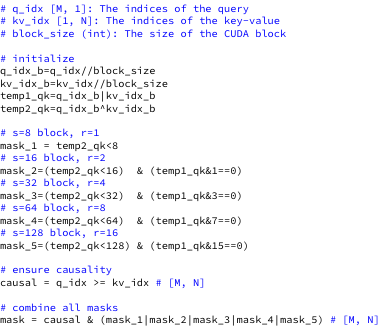}
        \end{algorithm}
    \end{minipage}
\end{figure}

%% file: main.bbl
\begin{thebibliography}{54}
\providecommand{\natexlab}[1]{#1}
\providecommand{\url}[1]{\texttt{#1}}
\expandafter\ifx\csname urlstyle\endcsname\relax
  \providecommand{\doi}[1]{doi: #1}\else
  \providecommand{\doi}{doi: \begingroup \urlstyle{rm}\Url}\fi

\bibitem[An et~al.(2023)An, Gong, Zhong, Li, Zhang, Kong, and Qiu]{leval}
An, C., Gong, S., Zhong, M., Li, M., Zhang, J., Kong, L., and Qiu, X.
\newblock L-eval: Instituting standardized evaluation for long context language models.
\newblock \emph{arXiv preprint arXiv:2307.11088}, 2023.

\bibitem[Anagnostidis et~al.(2023)Anagnostidis, Pavllo, Biggio, Noci, Lucchi, and Hofmann]{DBLP:conf/nips/AnagnostidisPBN23}
Anagnostidis, S., Pavllo, D., Biggio, L., Noci, L., Lucchi, A., and Hofmann, T.
\newblock Dynamic context pruning for efficient and interpretable autoregressive transformers.
\newblock In Oh, A., Naumann, T., Globerson, A., Saenko, K., Hardt, M., and Levine, S. (eds.), \emph{Advances in Neural Information Processing Systems 36: Annual Conference on Neural Information Processing Systems 2023, NeurIPS 2023, New Orleans, LA, USA, December 10 - 16, 2023}, 2023.

\bibitem[Arora et~al.(2024)Arora, Eyuboglu, Zhang, Timalsina, Alberti, Zou, Rudra, and R{\'{e}}]{DBLP:conf/icml/AroraEZTA0RR24}
Arora, S., Eyuboglu, S., Zhang, M., Timalsina, A., Alberti, S., Zou, J., Rudra, A., and R{\'{e}}, C.
\newblock Simple linear attention language models balance the recall-throughput tradeoff.
\newblock In \emph{Forty-first International Conference on Machine Learning, {ICML} 2024, Vienna, Austria, July 21-27, 2024}, 2024.

\bibitem[Behrouz et~al.(2024)Behrouz, Zhong, and Mirrokni]{behrouz2024titanslearningmemorizetest}
Behrouz, A., Zhong, P., and Mirrokni, V.
\newblock Titans: Learning to memorize at test time, 2024.
\newblock URL \url{https://arxiv.org/abs/2501.00663}.

\bibitem[Beltagy et~al.(2020)Beltagy, Peters, and Cohan]{beltagy2020longformerlongdocumenttransformer}
Beltagy, I., Peters, M.~E., and Cohan, A.
\newblock Longformer: The long-document transformer, 2020.
\newblock URL \url{https://arxiv.org/abs/2004.05150}.

\bibitem[Chen et~al.(2023)Chen, Yuan, Ye, Majumder, and Richardson]{chen2023put}
Chen, J., Yuan, S., Ye, R., Majumder, B.~P., and Richardson, K.
\newblock Put your money where your mouth is: Evaluating strategic planning and execution of llm agents in an auction arena.
\newblock \emph{arXiv preprint arXiv:2310.05746}, 2023.

\bibitem[Chevalier et~al.(2024)Chevalier, Geng, Wettig, Chen, Mizera, Annala, Aragon, Fanlo, Frieder, Machado, et~al.]{chevalier2024language}
Chevalier, A., Geng, J., Wettig, A., Chen, H., Mizera, S., Annala, T., Aragon, M.~J., Fanlo, A.~R., Frieder, S., Machado, S., et~al.
\newblock Language models as science tutors.
\newblock \emph{arXiv preprint arXiv:2402.11111}, 2024.

\bibitem[Child et~al.(2019)Child, Gray, Radford, and Sutskever]{child2019generatinglongsequencessparse}
Child, R., Gray, S., Radford, A., and Sutskever, I.
\newblock Generating long sequences with sparse transformers, 2019.
\newblock URL \url{https://arxiv.org/abs/1904.10509}.

\bibitem[Correia et~al.(2019)Correia, Niculae, and Martins]{DBLP:conf/emnlp/CorreiaNM19}
Correia, G.~M., Niculae, V., and Martins, A. F.~T.
\newblock Adaptively sparse transformers.
\newblock In Inui, K., Jiang, J., Ng, V., and Wan, X. (eds.), \emph{Proceedings of the 2019 Conference on Empirical Methods in Natural Language Processing and the 9th International Joint Conference on Natural Language Processing, {EMNLP-IJCNLP} 2019, Hong Kong, China, November 3-7, 2019}, pp.\  2174--2184. Association for Computational Linguistics, 2019.
\newblock \doi{10.18653/V1/D19-1223}.
\newblock URL \url{https://doi.org/10.18653/v1/D19-1223}.

\bibitem[De et~al.(2024)De, Smith, Fernando, Botev, Cristian-Muraru, Gu, Haroun, Berrada, Chen, Srinivasan, Desjardins, Doucet, Budden, Teh, Pascanu, Freitas, and Gulcehre]{de2024griffinmixinggatedlinear}
De, S., Smith, S.~L., Fernando, A., Botev, A., Cristian-Muraru, G., Gu, A., Haroun, R., Berrada, L., Chen, Y., Srinivasan, S., Desjardins, G., Doucet, A., Budden, D., Teh, Y.~W., Pascanu, R., Freitas, N.~D., and Gulcehre, C.
\newblock Griffin: Mixing gated linear recurrences with local attention for efficient language models, 2024.
\newblock URL \url{https://arxiv.org/abs/2402.19427}.

\bibitem[DeepSeek-AI et~al.(2025)DeepSeek-AI, Guo, Yang, Zhang, Song, Zhang, Xu, Zhu, Ma, Wang, Bi, Zhang, Yu, Wu, Wu, Gou, Shao, Li, Gao, Liu, Xue, Wang, Wu, Feng, Lu, Zhao, Deng, Zhang, Ruan, Dai, Chen, Ji, Li, Lin, Dai, Luo, Hao, Chen, Li, Zhang, Bao, Xu, Wang, Ding, Xin, Gao, Qu, Li, Guo, Li, Wang, Chen, Yuan, Qiu, Li, Cai, Ni, Liang, Chen, Dong, Hu, Gao, Guan, Huang, Yu, Wang, Zhang, Zhao, Wang, Zhang, Xu, Xia, Zhang, Zhang, Tang, Li, Wang, Li, Tian, Huang, Zhang, Wang, Chen, Du, Ge, Zhang, Pan, Wang, Chen, Jin, Chen, Lu, Zhou, Chen, Ye, Wang, Yu, Zhou, Pan, Li, Zhou, Wu, Ye, Yun, Pei, Sun, Wang, Zeng, Zhao, Liu, Liang, Gao, Yu, Zhang, Xiao, An, Liu, Wang, Chen, Nie, Cheng, Liu, Xie, Liu, Yang, Li, Su, Lin, Li, Jin, Shen, Chen, Sun, Wang, Song, Zhou, Wang, Shan, Li, Wang, Wei, Zhang, Xu, Li, Zhao, Sun, Wang, Yu, Zhang, Shi, Xiong, He, Piao, Wang, Tan, Ma, Liu, Guo, Ou, Wang, Gong, Zou, He, Xiong, Luo, You, Liu, Zhou, Zhu, Xu, Huang, Li, Zheng, Zhu, Ma, Tang, Zha, Yan, Ren, Ren, Sha, Fu, Xu, Xie, Zhang,
  Hao, Ma, Yan, Wu, Gu, Zhu, Liu, Li, Xie, Song, Pan, Huang, Xu, Zhang, and Zhang]{deepseekai2025deepseekr1incentivizingreasoningcapability}
DeepSeek-AI, Guo, D., Yang, D., Zhang, H., Song, J., Zhang, R., Xu, R., Zhu, Q., Ma, S., Wang, P., Bi, X., Zhang, X., Yu, X., Wu, Y., Wu, Z.~F., Gou, Z., Shao, Z., Li, Z., Gao, Z., Liu, A., Xue, B., Wang, B., Wu, B., Feng, B., Lu, C., Zhao, C., Deng, C., Zhang, C., Ruan, C., Dai, D., Chen, D., Ji, D., Li, E., Lin, F., Dai, F., Luo, F., Hao, G., Chen, G., Li, G., Zhang, H., Bao, H., Xu, H., Wang, H., Ding, H., Xin, H., Gao, H., Qu, H., Li, H., Guo, J., Li, J., Wang, J., Chen, J., Yuan, J., Qiu, J., Li, J., Cai, J.~L., Ni, J., Liang, J., Chen, J., Dong, K., Hu, K., Gao, K., Guan, K., Huang, K., Yu, K., Wang, L., Zhang, L., Zhao, L., Wang, L., Zhang, L., Xu, L., Xia, L., Zhang, M., Zhang, M., Tang, M., Li, M., Wang, M., Li, M., Tian, N., Huang, P., Zhang, P., Wang, Q., Chen, Q., Du, Q., Ge, R., Zhang, R., Pan, R., Wang, R., Chen, R.~J., Jin, R.~L., Chen, R., Lu, S., Zhou, S., Chen, S., Ye, S., Wang, S., Yu, S., Zhou, S., Pan, S., Li, S.~S., Zhou, S., Wu, S., Ye, S., Yun, T., Pei, T., Sun, T., Wang, T., Zeng, W.,
  Zhao, W., Liu, W., Liang, W., Gao, W., Yu, W., Zhang, W., Xiao, W.~L., An, W., Liu, X., Wang, X., Chen, X., Nie, X., Cheng, X., Liu, X., Xie, X., Liu, X., Yang, X., Li, X., Su, X., Lin, X., Li, X.~Q., Jin, X., Shen, X., Chen, X., Sun, X., Wang, X., Song, X., Zhou, X., Wang, X., Shan, X., Li, Y.~K., Wang, Y.~Q., Wei, Y.~X., Zhang, Y., Xu, Y., Li, Y., Zhao, Y., Sun, Y., Wang, Y., Yu, Y., Zhang, Y., Shi, Y., Xiong, Y., He, Y., Piao, Y., Wang, Y., Tan, Y., Ma, Y., Liu, Y., Guo, Y., Ou, Y., Wang, Y., Gong, Y., Zou, Y., He, Y., Xiong, Y., Luo, Y., You, Y., Liu, Y., Zhou, Y., Zhu, Y.~X., Xu, Y., Huang, Y., Li, Y., Zheng, Y., Zhu, Y., Ma, Y., Tang, Y., Zha, Y., Yan, Y., Ren, Z.~Z., Ren, Z., Sha, Z., Fu, Z., Xu, Z., Xie, Z., Zhang, Z., Hao, Z., Ma, Z., Yan, Z., Wu, Z., Gu, Z., Zhu, Z., Liu, Z., Li, Z., Xie, Z., Song, Z., Pan, Z., Huang, Z., Xu, Z., Zhang, Z., and Zhang, Z.
\newblock Deepseek-r1: Incentivizing reasoning capability in llms via reinforcement learning, 2025.
\newblock URL \url{https://arxiv.org/abs/2501.12948}.

\bibitem[Ding et~al.(2023)Ding, Ma, Dong, Zhang, Huang, Wang, Zheng, and Wei]{ding2023longnetscalingtransformers1000000000}
Ding, J., Ma, S., Dong, L., Zhang, X., Huang, S., Wang, W., Zheng, N., and Wei, F.
\newblock Longnet: Scaling transformers to 1,000,000,000 tokens, 2023.
\newblock URL \url{https://arxiv.org/abs/2307.02486}.

\bibitem[Dong et~al.(2024)Dong, Feng, Guessous, Liang, and He]{dong2024flex}
Dong, J., Feng, B., Guessous, D., Liang, Y., and He, H.
\newblock Flex attention: A programming model for generating optimized attention kernels.
\newblock \emph{arXiv preprint arXiv:2412.05496}, 2024.

\bibitem[Feng et~al.(2024)Feng, Tung, Hajimirsadeghi, Ahmed, Bengio, and Mori]{feng2024attentionrnn}
Feng, L., Tung, F., Hajimirsadeghi, H., Ahmed, M.~O., Bengio, Y., and Mori, G.
\newblock Attention as an rnn, 2024.
\newblock URL \url{https://arxiv.org/abs/2405.13956}.

\bibitem[Gao et~al.(2024)Gao, Zeng, Du, Cao, So, Cao, Yang, and Yang]{gao2024seerattentionlearningintrinsicsparse}
Gao, Y., Zeng, Z., Du, D., Cao, S., So, H. K.-H., Cao, T., Yang, F., and Yang, M.
\newblock Seerattention: Learning intrinsic sparse attention in your llms, 2024.
\newblock URL \url{https://arxiv.org/abs/2410.13276}.

\bibitem[Ge et~al.(2024)Ge, Zhang, Liu, Zhang, Han, and Gao]{DBLP:conf/iclr/Ge0LZ0024}
Ge, S., Zhang, Y., Liu, L., Zhang, M., Han, J., and Gao, J.
\newblock Model tells you what to discard: Adaptive {KV} cache compression for llms.
\newblock In \emph{The Twelfth International Conference on Learning Representations, {ICLR} 2024, Vienna, Austria, May 7-11, 2024}, 2024.

\bibitem[Gu \& Dao(2024)Gu and Dao]{gu2024mambalineartimesequencemodeling}
Gu, A. and Dao, T.
\newblock Mamba: Linear-time sequence modeling with selective state spaces, 2024.
\newblock URL \url{https://arxiv.org/abs/2312.00752}.

\bibitem[Han et~al.(2024)Han, Wang, Peng, Xiong, Chen, Ji, and Wang]{DBLP:conf/naacl/HanWPX0JW24}
Han, C., Wang, Q., Peng, H., Xiong, W., Chen, Y., Ji, H., and Wang, S.
\newblock Lm-infinite: Zero-shot extreme length generalization for large language models.
\newblock In Duh, K., G{\'{o}}mez{-}Adorno, H., and Bethard, S. (eds.), \emph{Proceedings of the 2024 Conference of the North American Chapter of the Association for Computational Linguistics: Human Language Technologies (Volume 1: Long Papers), {NAACL} 2024, Mexico City, Mexico, June 16-21, 2024}, pp.\  3991--4008. Association for Computational Linguistics, 2024.
\newblock \doi{10.18653/V1/2024.NAACL-LONG.222}.
\newblock URL \url{https://doi.org/10.18653/v1/2024.naacl-long.222}.

\bibitem[Hooper et~al.(2024)Hooper, Kim, Mohammadzadeh, Maheswaran, Paik, Mahoney, Keutzer, and Gholami]{hooper2024squeezedattentionacceleratinglong}
Hooper, C., Kim, S., Mohammadzadeh, H., Maheswaran, M., Paik, J., Mahoney, M.~W., Keutzer, K., and Gholami, A.
\newblock Squeezed attention: Accelerating long context length llm inference, 2024.
\newblock URL \url{https://arxiv.org/abs/2411.09688}.

\bibitem[Hsieh et~al.(2024)Hsieh, Sun, Kriman, Acharya, Rekesh, Jia, Zhang, and Ginsburg]{hsieh2024rulerwhatsrealcontext}
Hsieh, C.-P., Sun, S., Kriman, S., Acharya, S., Rekesh, D., Jia, F., Zhang, Y., and Ginsburg, B.
\newblock Ruler: What's the real context size of your long-context language models?, 2024.
\newblock URL \url{https://arxiv.org/abs/2404.06654}.

\bibitem[Jiang et~al.(2024)Jiang, LI, Zhang, Wu, Luo, Ahn, Han, Abdi, Li, Lin, et~al.]{jiangminference}
Jiang, H., LI, Y., Zhang, C., Wu, Q., Luo, X., Ahn, S., Han, Z., Abdi, A.~H., Li, D., Lin, C.-Y., et~al.
\newblock Minference 1.0: Accelerating pre-filling for long-context llms via dynamic sparse attention.
\newblock In \emph{The Thirty-eighth Annual Conference on Neural Information Processing Systems}, 2024.

\bibitem[Katharopoulos et~al.(2020)Katharopoulos, Vyas, Pappas, and Fleuret]{DBLP:conf/icml/KatharopoulosV020}
Katharopoulos, A., Vyas, A., Pappas, N., and Fleuret, F.
\newblock Transformers are rnns: Fast autoregressive transformers with linear attention.
\newblock In \emph{Proceedings of the 37th International Conference on Machine Learning, {ICML} 2020, 13-18 July 2020, Virtual Event}, volume 119 of \emph{Proceedings of Machine Learning Research}, pp.\  5156--5165. {PMLR}, 2020.

\bibitem[Kim et~al.(2024)Kim, Shim, Choi, and Chang]{DBLP:conf/emnlp/KimSCC24}
Kim, M., Shim, K., Choi, J., and Chang, S.
\newblock Infinipot: Infinite context processing on memory-constrained llms.
\newblock In Al{-}Onaizan, Y., Bansal, M., and Chen, Y. (eds.), \emph{Proceedings of the 2024 Conference on Empirical Methods in Natural Language Processing, {EMNLP} 2024, Miami, FL, USA, November 12-16, 2024}, pp.\  16046--16060. Association for Computational Linguistics, 2024.
\newblock URL \url{https://aclanthology.org/2024.emnlp-main.897}.

\bibitem[Li et~al.(2023)Li, Cai, Zhang, Chen, and Dey]{DBLP:conf/iclr/LiCZCD23}
Li, Y., Cai, T., Zhang, Y., Chen, D., and Dey, D.
\newblock What makes convolutional models great on long sequence modeling?
\newblock In \emph{The Eleventh International Conference on Learning Representations, {ICLR} 2023, Kigali, Rwanda, May 1-5, 2023}, 2023.

\bibitem[Li et~al.(2024)Li, Huang, Yang, Venkitesh, Locatelli, Ye, Cai, Lewis, and Chen]{li2024snapkvllmknowslooking}
Li, Y., Huang, Y., Yang, B., Venkitesh, B., Locatelli, A., Ye, H., Cai, T., Lewis, P., and Chen, D.
\newblock Snapkv: Llm knows what you are looking for before generation, 2024.
\newblock URL \url{https://arxiv.org/abs/2404.14469}.

\bibitem[Lieber et~al.(2024)Lieber, Lenz, Bata, Cohen, Osin, Dalmedigos, Safahi, Meirom, Belinkov, Shalev-Shwartz, Abend, Alon, Asida, Bergman, Glozman, Gokhman, Manevich, Ratner, Rozen, Shwartz, Zusman, and Shoham]{lieber2024jambahybridtransformermambalanguage}
Lieber, O., Lenz, B., Bata, H., Cohen, G., Osin, J., Dalmedigos, I., Safahi, E., Meirom, S., Belinkov, Y., Shalev-Shwartz, S., Abend, O., Alon, R., Asida, T., Bergman, A., Glozman, R., Gokhman, M., Manevich, A., Ratner, N., Rozen, N., Shwartz, E., Zusman, M., and Shoham, Y.
\newblock Jamba: A hybrid transformer-mamba language model, 2024.
\newblock URL \url{https://arxiv.org/abs/2403.19887}.

\bibitem[Liu et~al.(2024)Liu, Zaharia, and Abbeel]{liu2024ringattention}
Liu, H., Zaharia, M., and Abbeel, P.
\newblock Ringattention with blockwise transformers for near-infinite context.
\newblock In \emph{The Twelfth International Conference on Learning Representations}, 2024.
\newblock URL \url{https://openreview.net/forum?id=WsRHpHH4s0}.

\bibitem[Liu et~al.(2022)Liu, Qu, Chen, Tu, Ding, and Xie]{9896137}
Liu, L., Qu, Z., Chen, Z., Tu, F., Ding, Y., and Xie, Y.
\newblock Dynamic sparse attention for scalable transformer acceleration.
\newblock \emph{IEEE Transactions on Computers}, 71\penalty0 (12):\penalty0 3165--3178, 2022.
\newblock \doi{10.1109/TC.2022.3208206}.

\bibitem[Mohtashami \& Jaggi(2023)Mohtashami and Jaggi]{mohtashami2023randomaccess}
Mohtashami, A. and Jaggi, M.
\newblock Random-access infinite context length for transformers.
\newblock In \emph{Thirty-seventh Conference on Neural Information Processing Systems}, 2023.
\newblock URL \url{https://openreview.net/forum?id=7eHn64wOVy}.

\bibitem[OpenAI(2024)]{openai2024o1}
OpenAI.
\newblock Openai o1, 2024.
\newblock URL \url{https://openai.com/index/learning-to-reason-with-llms/}.

\bibitem[Park et~al.(2023)Park, O'Brien, Cai, Morris, Liang, and Bernstein]{Park2023GenerativeAgents}
Park, J.~S., O'Brien, J.~C., Cai, C.~J., Morris, M.~R., Liang, P., and Bernstein, M.~S.
\newblock Generative agents: Interactive simulacra of human behavior.
\newblock In \emph{In the 36th Annual ACM Symposium on User Interface Software and Technology (UIST '23)}, UIST '23, New York, NY, USA, 2023. Association for Computing Machinery.

\bibitem[Peng et~al.(2023)Peng, Alcaide, Anthony, Albalak, Arcadinho, Biderman, Cao, Cheng, Chung, Grella, GV, He, Hou, Lin, Kazienko, Kocon, Kong, Koptyra, Lau, Mantri, Mom, Saito, Song, Tang, Wang, Wind, Wozniak, Zhang, Zhang, Zhao, Zhou, Zhou, Zhu, and Zhu]{peng2023rwkvreinventingrnnstransformer}
Peng, B., Alcaide, E., Anthony, Q., Albalak, A., Arcadinho, S., Biderman, S., Cao, H., Cheng, X., Chung, M., Grella, M., GV, K.~K., He, X., Hou, H., Lin, J., Kazienko, P., Kocon, J., Kong, J., Koptyra, B., Lau, H., Mantri, K. S.~I., Mom, F., Saito, A., Song, G., Tang, X., Wang, B., Wind, J.~S., Wozniak, S., Zhang, R., Zhang, Z., Zhao, Q., Zhou, P., Zhou, Q., Zhu, J., and Zhu, R.-J.
\newblock Rwkv: Reinventing rnns for the transformer era, 2023.
\newblock URL \url{https://arxiv.org/abs/2305.13048}.

\bibitem[Poli et~al.(2023{\natexlab{a}})Poli, Massaroli, Nguyen, Fu, Dao, Baccus, Bengio, Ermon, and R{\'{e}}]{DBLP:conf/icml/PoliMNFDBBER23}
Poli, M., Massaroli, S., Nguyen, E., Fu, D.~Y., Dao, T., Baccus, S., Bengio, Y., Ermon, S., and R{\'{e}}, C.
\newblock Hyena hierarchy: Towards larger convolutional language models.
\newblock In Krause, A., Brunskill, E., Cho, K., Engelhardt, B., Sabato, S., and Scarlett, J. (eds.), \emph{International Conference on Machine Learning, {ICML} 2023, 23-29 July 2023, Honolulu, Hawaii, {USA}}, volume 202 of \emph{Proceedings of Machine Learning Research}, pp.\  28043--28078. {PMLR}, 2023{\natexlab{a}}.

\bibitem[Poli et~al.(2023{\natexlab{b}})Poli, Wang, Massaroli, Quesnelle, Carlow, Nguyen, and Thomas]{stripedhyena}
Poli, M., Wang, J., Massaroli, S., Quesnelle, J., Carlow, R., Nguyen, E., and Thomas, A.
\newblock {StripedHyena: Moving Beyond Transformers with Hybrid Signal Processing Models}, 12 2023{\natexlab{b}}.
\newblock URL \url{https://github.com/togethercomputer/stripedhyena}.

\bibitem[Qu et~al.(2022)Qu, Liu, Tu, Chen, Ding, and Xie]{DBLP:conf/asplos/Qu0TCD022}
Qu, Z., Liu, L., Tu, F., Chen, Z., Ding, Y., and Xie, Y.
\newblock {DOTA:} detect and omit weak attentions for scalable transformer acceleration.
\newblock In Falsafi, B., Ferdman, M., Lu, S., and Wenisch, T.~F. (eds.), \emph{{ASPLOS} '22: 27th {ACM} International Conference on Architectural Support for Programming Languages and Operating Systems, Lausanne, Switzerland, 28 February 2022 - 4 March 2022}, pp.\  14--26. {ACM}, 2022.
\newblock \doi{10.1145/3503222.3507738}.
\newblock URL \url{https://doi.org/10.1145/3503222.3507738}.

\bibitem[Qwen(2024)]{qwq2024}
Qwen.
\newblock Qwq, 2024.
\newblock URL \url{https://qwenlm.github.io/blog/qwq-32b-preview/}.

\bibitem[Rae et~al.(2019)Rae, Potapenko, Jayakumar, Hillier, and Lillicrap]{raecompressive2019}
Rae, J.~W., Potapenko, A., Jayakumar, S.~M., Hillier, C., and Lillicrap, T.~P.
\newblock Compressive transformers for long-range sequence modelling.
\newblock \emph{arXiv preprint}, 2019.
\newblock URL \url{https://arxiv.org/abs/1911.05507}.

\bibitem[Ribar et~al.(2024)Ribar, Chelombiev, Hudlass{-}Galley, Blake, Luschi, and Orr]{DBLP:conf/icml/RibarCHBLO24}
Ribar, L., Chelombiev, I., Hudlass{-}Galley, L., Blake, C., Luschi, C., and Orr, D.
\newblock Sparq attention: Bandwidth-efficient {LLM} inference.
\newblock In \emph{Forty-first International Conference on Machine Learning, {ICML} 2024, Vienna, Austria, July 21-27, 2024}, 2024.

\bibitem[Roy et~al.(2021)Roy, Saffar, Vaswani, and Grangier]{DBLP:journals/tacl/RoySVG21}
Roy, A., Saffar, M., Vaswani, A., and Grangier, D.
\newblock Efficient content-based sparse attention with routing transformers.
\newblock \emph{Trans. Assoc. Comput. Linguistics}, 9:\penalty0 53--68, 2021.
\newblock \doi{10.1162/TACL_A_00353}.
\newblock URL \url{https://doi.org/10.1162/tacl\_a\_00353}.

\bibitem[Shi et~al.(2021)Shi, Gao, Ren, Xu, Liang, Li, and Kwok]{DBLP:conf/icml/ShiGRXLLK21}
Shi, H., Gao, J., Ren, X., Xu, H., Liang, X., Li, Z., and Kwok, J.~T.
\newblock Sparsebert: Rethinking the importance analysis in self-attention.
\newblock In Meila, M. and Zhang, T. (eds.), \emph{Proceedings of the 38th International Conference on Machine Learning, {ICML} 2021, 18-24 July 2021, Virtual Event}, volume 139 of \emph{Proceedings of Machine Learning Research}, pp.\  9547--9557. {PMLR}, 2021.
\newblock URL \url{http://proceedings.mlr.press/v139/shi21a.html}.

\bibitem[Soboleva et~al.(2023)Soboleva, Al-Khateeb, Myers, Steeves, Hestness, and Dey]{cerebras2023slimpajama}
Soboleva, D., Al-Khateeb, F., Myers, R., Steeves, J.~R., Hestness, J., and Dey, N.
\newblock {SlimPajama: A 627B token cleaned and deduplicated version of RedPajama}.
\newblock \url{https://cerebras.ai/blog/slimpajama-a-627b-token-cleaned-and-deduplicated-version-of-redpajama}, 2023.
\newblock URL \url{https://huggingface.co/datasets/cerebras/SlimPajama-627B}.

\bibitem[Sun et~al.(2023)Sun, Dong, Huang, Ma, Xia, Xue, Wang, and Wei]{sun2023retentivenetworksuccessortransformer}
Sun, Y., Dong, L., Huang, S., Ma, S., Xia, Y., Xue, J., Wang, J., and Wei, F.
\newblock Retentive network: A successor to transformer for large language models, 2023.
\newblock URL \url{https://arxiv.org/abs/2307.08621}.

\bibitem[Tang et~al.(2024)Tang, Zhao, Zhu, Xiao, Kasikci, and Han]{DBLP:conf/icml/TangZZXKH24}
Tang, J., Zhao, Y., Zhu, K., Xiao, G., Kasikci, B., and Han, S.
\newblock {QUEST:} query-aware sparsity for efficient long-context {LLM} inference.
\newblock In \emph{Forty-first International Conference on Machine Learning, {ICML} 2024, Vienna, Austria, July 21-27, 2024}, 2024.

\bibitem[Tillet et~al.(2019)Tillet, Kung, and Cox]{tillet2019triton}
Tillet, P., Kung, H.-T., and Cox, D.
\newblock Triton: an intermediate language and compiler for tiled neural network computations.
\newblock In \emph{Proceedings of the 3rd ACM SIGPLAN International Workshop on Machine Learning and Programming Languages}, pp.\  10--19, 2019.

\bibitem[Wang et~al.(2024)Wang, Chen, Cheng, Liao, Zhang, Wu, Yu, Xu, Zhang, Luo, Li, Yang, Huang, and Li]{wang-etal-2024-leave}
Wang, M., Chen, L., Cheng, F., Liao, S., Zhang, X., Wu, B., Yu, H., Xu, N., Zhang, L., Luo, R., Li, Y., Yang, M., Huang, F., and Li, Y.
\newblock Leave no document behind: Benchmarking long-context {LLM}s with extended multi-doc {QA}.
\newblock In Al-Onaizan, Y., Bansal, M., and Chen, Y.-N. (eds.), \emph{Proceedings of the 2024 Conference on Empirical Methods in Natural Language Processing}, pp.\  5627--5646, Miami, Florida, USA, November 2024. Association for Computational Linguistics.
\newblock \doi{10.18653/v1/2024.emnlp-main.322}.
\newblock URL \url{https://aclanthology.org/2024.emnlp-main.322/}.

\bibitem[Xiao et~al.(2024)Xiao, Tian, Chen, Han, and Lewis]{DBLP:conf/iclr/XiaoTCHL24}
Xiao, G., Tian, Y., Chen, B., Han, S., and Lewis, M.
\newblock Efficient streaming language models with attention sinks.
\newblock In \emph{The Twelfth International Conference on Learning Representations, {ICLR} 2024, Vienna, Austria, May 7-11, 2024}, 2024.

\bibitem[Xu et~al.(2024)Xu, Ping, Wu, Xu, Liu, Shoeybi, and Catanzaro]{xu2024chatqa2bridginggap}
Xu, P., Ping, W., Wu, X., Xu, C., Liu, Z., Shoeybi, M., and Catanzaro, B.
\newblock Chatqa 2: Bridging the gap to proprietary llms in long context and rag capabilities, 2024.
\newblock URL \url{https://arxiv.org/abs/2407.14482}.

\bibitem[Yang et~al.(2024{\natexlab{a}})Yang, Yang, Hui, Zheng, Yu, Zhou, Li, Li, Liu, Huang, Dong, Wei, Lin, Tang, Wang, Yang, Tu, Zhang, Ma, Yang, Xu, Zhou, Bai, He, Lin, Dang, Lu, Chen, Yang, Li, Xue, Ni, Zhang, Wang, Peng, Men, Gao, Lin, Wang, Bai, Tan, Zhu, Li, Liu, Ge, Deng, Zhou, Ren, Zhang, Wei, Ren, Liu, Fan, Yao, Zhang, Wan, Chu, Liu, Cui, Zhang, Guo, and Fan]{yang2024qwen2technicalreport}
Yang, A., Yang, B., Hui, B., Zheng, B., Yu, B., Zhou, C., Li, C., Li, C., Liu, D., Huang, F., Dong, G., Wei, H., Lin, H., Tang, J., Wang, J., Yang, J., Tu, J., Zhang, J., Ma, J., Yang, J., Xu, J., Zhou, J., Bai, J., He, J., Lin, J., Dang, K., Lu, K., Chen, K., Yang, K., Li, M., Xue, M., Ni, N., Zhang, P., Wang, P., Peng, R., Men, R., Gao, R., Lin, R., Wang, S., Bai, S., Tan, S., Zhu, T., Li, T., Liu, T., Ge, W., Deng, X., Zhou, X., Ren, X., Zhang, X., Wei, X., Ren, X., Liu, X., Fan, Y., Yao, Y., Zhang, Y., Wan, Y., Chu, Y., Liu, Y., Cui, Z., Zhang, Z., Guo, Z., and Fan, Z.
\newblock Qwen2 technical report, 2024{\natexlab{a}}.
\newblock URL \url{https://arxiv.org/abs/2407.10671}.

\bibitem[Yang et~al.(2024{\natexlab{b}})Yang, Zhang, Chen, Li, and Jia]{yang2024tidaldecodefastaccuratellm}
Yang, L., Zhang, Z., Chen, Z., Li, Z., and Jia, Z.
\newblock Tidaldecode: Fast and accurate llm decoding with position persistent sparse attention, 2024{\natexlab{b}}.
\newblock URL \url{https://arxiv.org/abs/2410.05076}.

\bibitem[Yang et~al.(2025)Yang, Wang, Zhang, Shen, and Kim]{yang2025parallelizinglineartransformersdelta}
Yang, S., Wang, B., Zhang, Y., Shen, Y., and Kim, Y.
\newblock Parallelizing linear transformers with the delta rule over sequence length, 2025.
\newblock URL \url{https://arxiv.org/abs/2406.06484}.

\bibitem[Zaheer et~al.(2020)Zaheer, Guruganesh, Dubey, Ainslie, Alberti, Onta{\~{n}}{\'{o}}n, Pham, Ravula, Wang, Yang, and Ahmed]{DBLP:conf/nips/ZaheerGDAAOPRWY20}
Zaheer, M., Guruganesh, G., Dubey, K.~A., Ainslie, J., Alberti, C., Onta{\~{n}}{\'{o}}n, S., Pham, P., Ravula, A., Wang, Q., Yang, L., and Ahmed, A.
\newblock Big bird: Transformers for longer sequences.
\newblock In Larochelle, H., Ranzato, M., Hadsell, R., Balcan, M., and Lin, H. (eds.), \emph{Advances in Neural Information Processing Systems 33: Annual Conference on Neural Information Processing Systems 2020, NeurIPS 2020, December 6-12, 2020, virtual}, 2020.

\bibitem[Zhang et~al.(2023)Zhang, Sheng, Zhou, Chen, Zheng, Cai, Song, Tian, R{\'{e}}, Barrett, Wang, and Chen]{DBLP:conf/nips/Zhang00CZC0TRBW23}
Zhang, Z., Sheng, Y., Zhou, T., Chen, T., Zheng, L., Cai, R., Song, Z., Tian, Y., R{\'{e}}, C., Barrett, C.~W., Wang, Z., and Chen, B.
\newblock {H2O:} heavy-hitter oracle for efficient generative inference of large language models.
\newblock In Oh, A., Naumann, T., Globerson, A., Saenko, K., Hardt, M., and Levine, S. (eds.), \emph{Advances in Neural Information Processing Systems 36: Annual Conference on Neural Information Processing Systems 2023, NeurIPS 2023, New Orleans, LA, USA, December 10 - 16, 2023}, 2023.

\bibitem[Zhang et~al.(2024)Zhang, Zhu, Yang, Xu, Li, Phothilimthana, and Jia]{DBLP:conf/icml/ZhangZYXLPJ24}
Zhang, Z., Zhu, A., Yang, L., Xu, Y., Li, L., Phothilimthana, P.~M., and Jia, Z.
\newblock Accelerating iterative retrieval-augmented language model serving with speculation.
\newblock In \emph{Forty-first International Conference on Machine Learning, {ICML} 2024, Vienna, Austria, July 21-27, 2024}, 2024.

\bibitem[Zhou et~al.(2023)Zhou, Xu, Zhu, Zhou, Lo, Sridhar, Cheng, Ou, Bisk, Fried, et~al.]{zhou2023webarena}
Zhou, S., Xu, F.~F., Zhu, H., Zhou, X., Lo, R., Sridhar, A., Cheng, X., Ou, T., Bisk, Y., Fried, D., et~al.
\newblock Webarena: A realistic web environment for building autonomous agents.
\newblock \emph{arXiv preprint arXiv:2307.13854}, 2023.

\end{thebibliography}
